\documentclass{article}

\usepackage{PRIMEarxiv}

\usepackage[utf8]{inputenc} 
\usepackage[T1]{fontenc}    
\usepackage{hyperref}       
\usepackage{url}            
\usepackage{booktabs}       
\usepackage{amsfonts}       
\usepackage{nicefrac}       
\usepackage{microtype}      
\usepackage{lipsum}
\usepackage{fancyhdr}       
\usepackage{graphicx}       
\graphicspath{{media/}}     
\usepackage{xcolor}
\usepackage{float}
\usepackage{natbib}
\definecolor{Gray}{gray}{0.6}
\usepackage{xcolor}
\usepackage{subcaption}
\title{Vector Quantization Loss Analysis in VQGANs: A Single-GPU Ablation Study for Image-to-Image Synthesis}

\author{Luv Verma\thanks{\textit{corresponding author}} \\
    Khoury College of Computer Sciences\\
    Northeastern University\\
    Boston, MA 02115 \\
    \texttt{verma.lu@northeastern.edu} \and 
    \textbf{Varun Mohan}\\
    Khoury College of Computer Sciences\\
    Northeastern University\\
    Boston, MA 02115 \\
    \texttt{mohan.va@northeastern.edu}
}

\begin{document}
\maketitle

\begin{abstract}
This study performs an ablation analysis of Vector Quantized Generative Adversarial Networks (VQGANs), concentrating on image-to-image synthesis utilizing a single NVIDIA A100 GPU. The current work explores the nuanced effects of varying critical parameters including the number of epochs, image count, and attributes of codebook vectors and latent dimensions, specifically within the constraint of limited resources. Notably, our focus is pinpointed on the vector quantization loss, keeping other hyperparameters and loss components (GAN loss) fixed. This was done to delve into a deeper understanding of the discrete latent space, and to explore how varying its size affects the reconstruction. Though, our results do not surpass the existing benchmarks, however, our findings shed significant light on VQGAN’s behaviour for a smaller dataset, particularly concerning artifacts, codebook size optimization, and comparative analysis with Principal Component Analysis (PCA). The study also uncovers the promising direction by introducing 2D positional encodings, revealing a marked reduction in artifacts and insights into balancing clarity and overfitting.

\end{abstract}


\section{Introduction}
The introduction of VQ-VAE models marked a transformative moment in the field of image processing, introducing the powerful concept of vector codebooks for discrete latent representation. VQ-VAEs, adept at modeling long-term dependencies, successfully harnessed the compressed discrete latent space to generate images, action sequences, and even meaningful speech in an unsupervised manner \citep{van2017neural}. Building on this foundation, VQGANs introduced a fusion of GANs for image reconstruction, with a particular emphasis on using transformer attention layers in both encoder and decoder \citep{esser2021taming}. This development has enabled high-resolution image synthesis through the innovative representation of images as compositions of perceptually rich constituents.
These advancements have showcased significant success stories, particularly in handling large image datasets, often surpassing state-of-the-art convolutional approaches \citep{esser2021taming}. The utilization of codebook vectors has not only reduced data requirements but also transitioned the modeling space from continuous to discrete.
In this study, we sought to examine how these sophisticated models behave when applied to a smaller, more constrained dataset with limited computational resources, such as a single GPU \(a100\) \cite{nvidia2020a100}. The Oxford 102 Flower dataset \cite{oxford102flowers}, with its rich variations in colors and features, serves as a suitable testing ground for this examination. By choosing subsets of images and varying codebook sizes and latent dimensions, this investigation methodically probes the impact of these parameters on image reconstruction. We are posing this work as a more of an ablation study where our specific focus is on the vector quantization process and therefore we have focused on evaluating vector quantization loss extensively, while keeping the GAN component fixed by not changing hyperparameters associated with it. 
Moreover, to provide a broader perspective, this study also includes an exploration of Principal Component Analysis (PCA) \cite{jolliffe2016principal} on the same dataset. This additional analysis serves as a comparative benchmark, offering insights into the relative strengths and challenges of different reconstruction techniques.
Taking motivation from NeRF-based models \cite{mildenhall2021nerf} and transformer-based models \cite{vaswani2017attention}, we also experimented with the image-to-image reconstruction process by adding positional encodings, an approach that opens doors to a deeper understanding of how different mathematical concepts can be combined to enhance reconstruction techniques.
By focusing on a scenario where abundant computational resources are not available, this exploration intends to contribute valuable insights into how the capabilities of VQ-VAE and VQGAN architectures can be leveraged and understood within such constraints. Such an investigation has the potential to inform future work, inspire new innovations, and facilitate more accessible research for those who may not have access to extensive resources.

\section{Details of the Image-to-Image Synthesis Model}
\label{sec:headings}


\subsection{Basic components}
\begin{figure}[!h]
\centering
\includegraphics[width=0.9\linewidth]{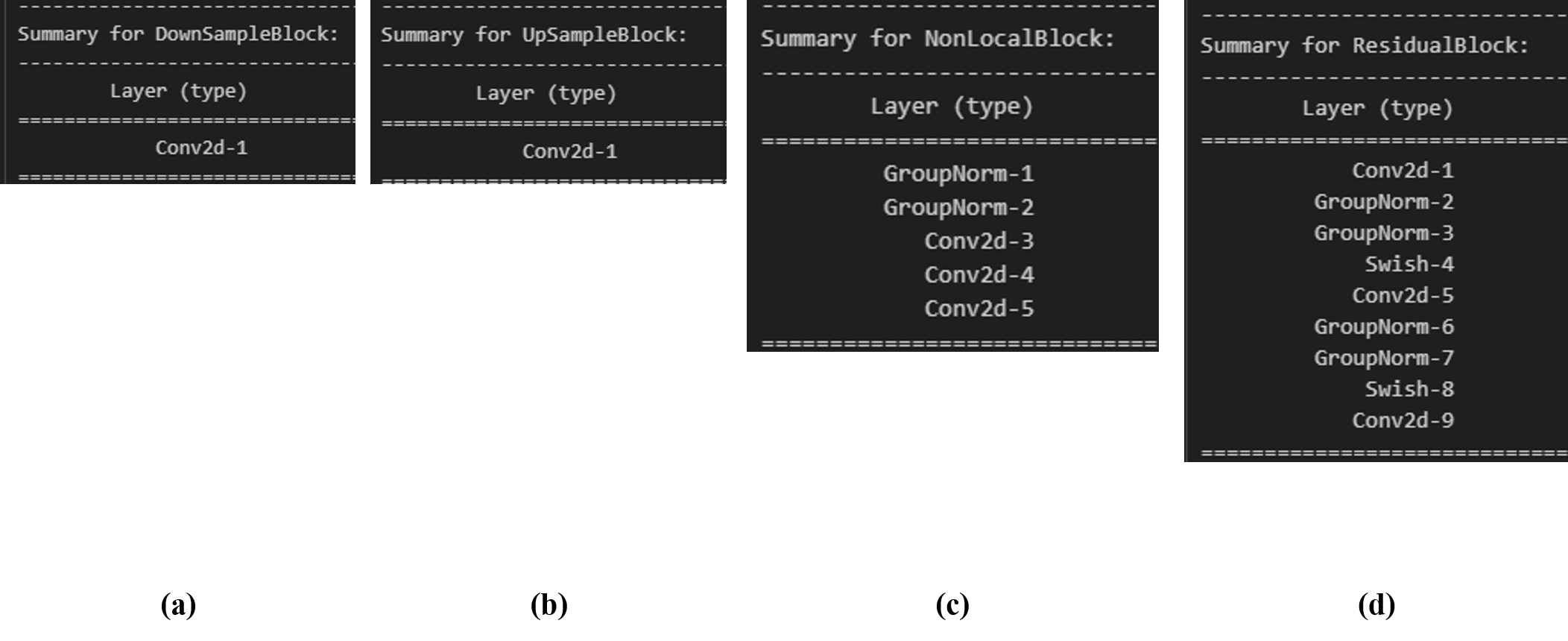}

\caption{\textbf{(a)} \textbf{Column 1}: Summary of DownSample Block. \textbf{(b)} \textbf{Column 2}: Summary of UpSample Block. \textbf{(c)} \textbf{Column 3}: Summary of NonLocal Block. \textbf{(d)} \textbf{Column 4}: Summary of Residual Block. }
\label{fig0}
\end{figure}

The code has been adopted and modified from VQGAN original implementation \citep{esser2021taming} and one more source from Github \citep{dome2722023vqgan}. With all the recent modifications, the code is available on GitHub\citep{luvverma2011}.
The basic components of the model included activation function, Residual Block, UpSample Block, DownSample Block and Non-Local Block. Their summaries are shown in figure  \ref{fig0}. Only the layers are mentioned, as number of filters, image height and width varied depending upon where these components are used. The Norm used was groupnorm and activation function used was swish activation. 

\begin{itemize}
\item \textbf{Swish Activation Function}: In comparison to more traditional functions like ReLU or Sigmoid, swish is known for its smooth gradient flow and non-zero-centered output.
\item \textbf{Residual Block}: 
Deep networks often struggle to learn effectively due to the diminishing magnitude of gradients during backpropagation, which leads to slower convergence and difficulty in optimizing the model's parameters. By using residual blocks, the VQGAN architecture can construct much deeper neural networks while maintaining efficient training. The skip connections in residual blocks allow the gradient to flow directly back to the earlier layers, bypassing several intermediate layers. This effectively mitigates the vanishing gradient issue and enables smoother and more effective training of the model.
\item \textbf{Up-Sample Block and Down-Sample Block}: These blocks are employed to manipulate the resolutions of the feature maps. The downsample block act as an encoder, and reduces the resolution to extract high-level features. The upsample block (equivalent to decoder layer) amplifies the resolution of the feature maps, making it essential for fine-grained reconstructions in our model. 

\item \textbf{Non-Local Block}: This block is designed to catch long-range connections within the input data. Typically, in a deep learning network, traditional layers (like convolutional layers) are limited to capturing only nearby or local information due to kernels. This approach might overlook some long-range connections across different parts of the input data. The non-local block helps overcome this limitation. It uses an attention mechanism that allows each feature in the data to interact with all other features. This means it can consider the global context of the data and can learn from patterns and connections that span across the entire input, which could lead to better performance on complex tasks.
\end{itemize}


\subsection{Encoder}
\begin{figure}[!h]
\centering

    \begin{subfigure}[b]{0.45\linewidth}
        \includegraphics[width=\linewidth]{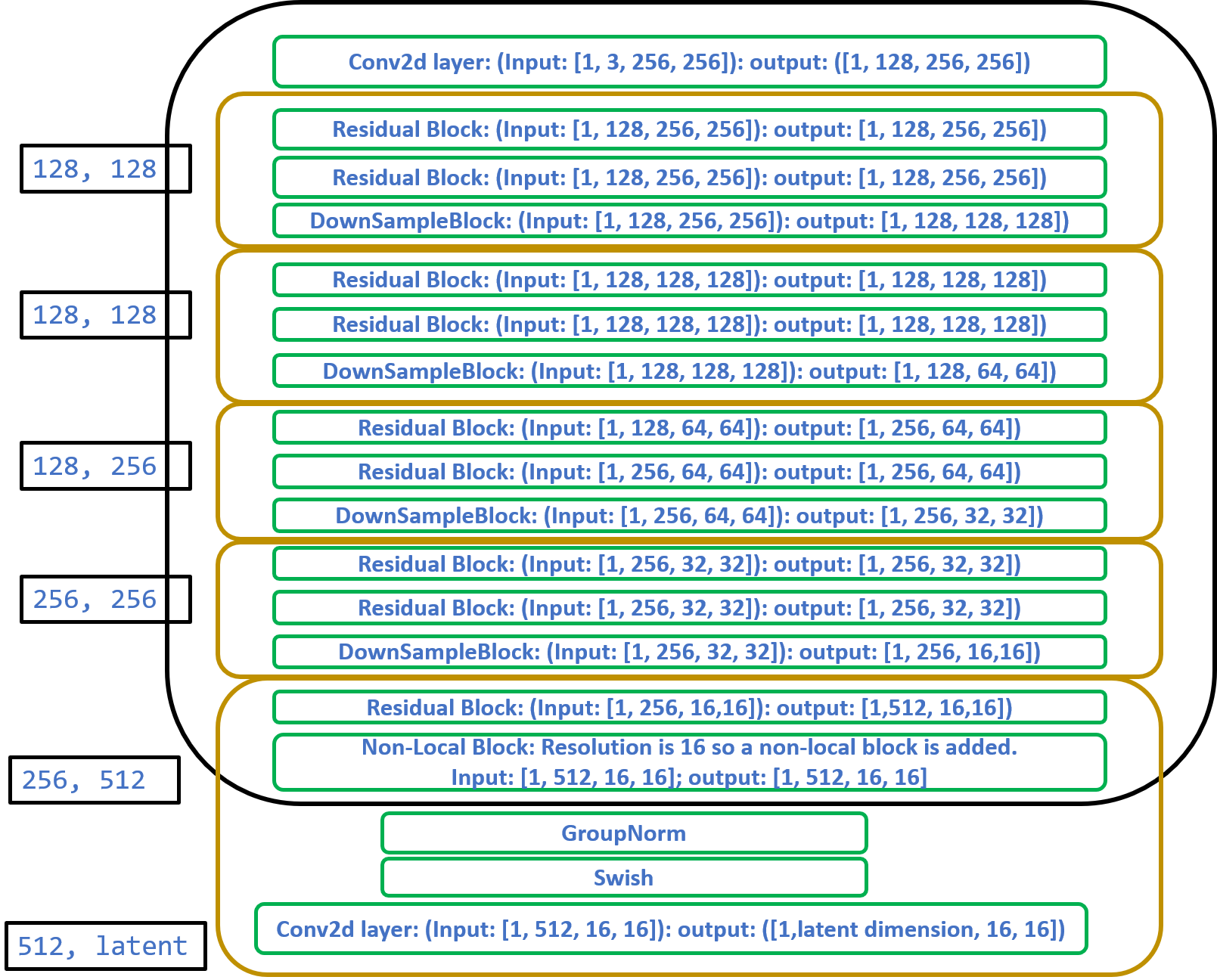}
        \caption{Encoder Layers}
        \label{fig:1a}
    \end{subfigure}
    \hfill
    \begin{subfigure}[b]{0.45\linewidth}
        \includegraphics[width=\linewidth]{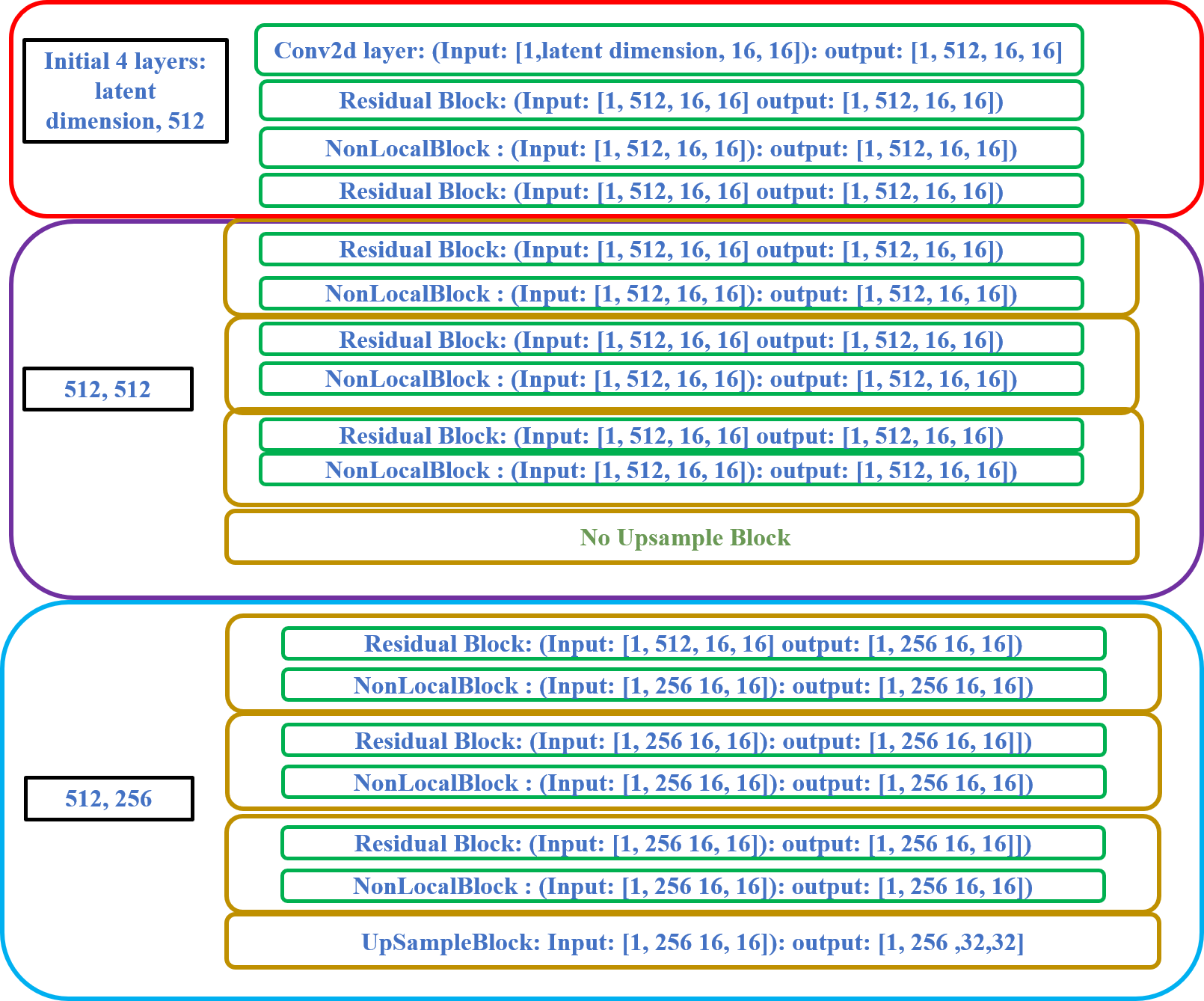}
        \caption{Decoder Layers}
        \label{fig:1b}
    \end{subfigure}
    \caption{Encoder and Decoder: Representation of the high level architecture. Refer to Figure~\ref{fig0} for details of each component}
    \label{fig1}
\end{figure}

As illustrated in figure \ref{fig1}, the Encoder takes high-dimensional image data as input and converts it into a compressed form that preserves the essential features necessary for the task. The first dimension in the figure, represented by no. 1, indicates the batch size which can vary. This compressed form, known as a latent representation, serves as a compact summary of the input data.

Refer to Figure~\ref{fig:1a} for part (a) and Figure~\ref{fig:1b} for part (b).

The Encoder accomplishes this through a series of transformations utilizing various components such as Convolutional layers, Residual Blocks, Non-Local Blocks, and DownSample Blocks and 2d positional encoding. These components work in tandem to extract and enhance the most significant features from the input while progressively reducing its dimensionality. Encoder layers are followed by decoder layers.

\subsection{Decoder}
The decoder module, serves a complimentary function to the encoder module, where it takes the low-dimensional latent representations generated by the encoder and transforms them back into high-dimensional output that mirrors the original input data, also known as reconstructed data. To achieve this, decoder uses a sequence of components such as convolutional layers, residual blocks, non-local blocks, upsample blocks and 2d positional encodings. 

The process begins with a convolutional layer and a positional encoding, followed by a series of residual blocks and non-local blocks, that serve to re-introduce the intricate features and long-range dependencies that were compressed in the encoding stage. UpSample blocks progressively increase the dimensionality of the representation to match that of the original data. another round of positional encoding is applied approximately in the middle of the network to further enhance the feature representation. The final stages of the decoder consist of a groupnorm and Swish activation function which outputs the reconstructed data, matching the channel dimensions of original input. 

\subsection{Codebook}

The Codebook serves as an essential component in Vector Quantization Generative Adversarial Networks (VQ-GANs) that facilitates the transformation of continuous latent representations into discrete form. The operational process of the Codebook can be summarized as follows:

\begin{itemize}

\item \textbf{Input Tensor:} The Codebook receives an input tensor with a shape (batchsize, latentdim, height, width), e.g., (5, 256, 16, 16).
    
\item \textbf{Input Reshaping:} The input tensor is reshaped to form a matrix of dimensions ($-1$, latentdim), e.g., (1280, 256). This transformation results in a set of 1024 vectors, each having 256 latent dimensions.

    \item \textbf{Vectors and Latent Dimensions:} We start with 1280 vectors, each of which possesses 256 latent dimensions.

    \item \textbf{Defining Embeddings:} For the embeddings, we define a matrix where each of the 256 latent dimensions can be represented by 1024 possible values. This is signified by the shape (1024, 256) of our embedding matrix.

    \item \textbf{Calculation of Euclidean Distance:} A computation of the Euclidean distance is performed between our reshaped input of (1280, 256) and our defined embeddings (256, 1024). This results in an output of shape (1280,1024). In simpler terms, for every one of our 1280 vectors, there exists one value (out of 1024 in the codebook) that has the minimum distance to the vector, serving as the nearest neighbor.
    
    \item \textbf{Identification of Nearest Neighbors:} Here, the index of the minimum value along the second dimension of our Euclidean distance matrix is identified. This essentially gives us the index of the nearest neighbor for every vector. Given that we have 1280 vectors, we end up with 1280 indices, each representing the nearest neighbor to a specific vector.

    \item \textbf{Mapping Vectors to Codebook Vectors:} With our collection of 1280 indices, we can determine which of our original vectors is closest to which codebook vector. In other words, for each latent dimension (out of 256) and a given vector (out of 1280), we can reference a specific codebook (out of 1024). If we expand this process to all the latent dimensions (256) for a specific vector (out of 1280), we choose the relevant codebook values (out of 1024). This means that for one vector, we have 256 values from the codebook.

    \item \textbf{Generation of Final Matrix:} By repeating the process detailed in step 5 for all 1280 vectors, we ultimately generate a matrix of size (1280, 256). In this matrix, the $256$ values for each of the 1280 vectors are derived from the codebook, providing a quantized representation of our original data.

\end{itemize}

The codebook serves as a dictionary for representing complex data more efficiently and accurately. It is critical because it enables a form of compression through quantization, helping to retain important information while reducing the size of the data, which is particularly useful in machine learning models for faster processing and lower memory requirements.

\subsection{Loss}

\begin{itemize}
\item \textbf{Reconstruction Loss:} This could be L1/L2 loss between the original image (input image) and the reconstructed image (output of the decoder, in which discrete latent representations are fed). 

\item \textbf{Commitment Loss:} This loss makes sure that the continuous representations ($z$) obtained from the encoder resembles the discrete representations (${z}_q$) which are obtained from the codebook. This is because, during inference, we just want to pass in the discrete representations (${z}_q$) to the decoder and generate a reasonable output image. Therefore, during training also decoder sees ${z}_q$ instead of ${z}$. As ${z}_q$ is derived from a non-differentiable operation (taking the argmin to select the closest codebook vector), it's detached from the computation graph, so that gradients dont flow through this operation. 

\item \textbf{Codebook Loss:} This part of the loss encourages the codebook to be used evenly. It tries to minimize the difference between continuous representation ($z$), which is detached from the graph in codebook loss and codebook vectors (${z}_q$). This helps in avoiding what are called as "dead" codebook vectors which are not close to any input vectors. 
One important thing to note about ${z}_q$ are that they are selected by argmin operation and are non-differentiable (i.e there gradients are non-computable), however, ${z}_q$ are the functions of codebook embeddings. Codebook embeddings are the weight of embedding layers (nn.Embedding in PyTorch), which are trainable and differentiable. Therefore, even though ${z}_q$ is detached from its selection operation (argmin), they are not detached from their codebook embeddings and thus their gradients have computed w.r.t codebook embedding parameters during the calculation of codebook loss.

\item \textbf{VQ Loss:} In the paper, VQ loss is given in equation 4 as a combination of three losses described above. 

\item \textbf{Perceptual Loss:} 
    \begin{itemize}
    \item \textbf{Introduction:} In the training process of the proposed VQGAN, a perceptual loss function is employed to ensure visual similarity between the generated and original images. This perceptual loss function guides the training such that the generated images resemble the original images not just in pixel space, but are also perceptually similar, making them more appealing to the human eye. It measures differences in a manner that closely aligns with human perceptual differences, contributing to the generation of higher quality images.
    
    \item \textbf{LPIPS Implementation:} The specific implementation of the perceptual loss function in this model is the Learnable Perceptual Image Patch Similarity (LPIPS). Here, the LPIPS metric is implemented as a PyTorch module. The LPIPS model relies on a pre-trained VGG16 network to extract feature representations from both real and generated images, and then measures the differences between these feature representations.
    
    \item \textbf{Preparation for Feature Extraction:} In the VQGAN model, both the real and generated images are first prepared for feature extraction by the pre-trained VGG16 network. Specifically, a module referred to as \texttt{ScalingLayer} is leveraged, which normalizes the RGB values of the images. The scaling and shifting of RGB values is crucial as it allows the VGG16 model to more effectively process and extract meaningful features from the images.
    
    \item \textbf{Feature Extraction and Normalization:} After this normalization process, both the real and generated images are passed to the VGG16 network. The VGG16 network serves as a feature extractor, breaking down the images into a set of 'feature maps' at various layers of the network. These feature maps are higher-level representations of the images, capturing more abstract visual characteristics compared to the raw pixel data. Once these feature maps are obtained, they are then normalized to become unit vectors via the norm tensor function. This normalization process, which generates unit vectors, ensures that the feature maps from different images can be compared on a common scale.
    
    \item \textbf{Difference Calculation and Final Remarks:} Subsequently, the squared difference between the corresponding feature maps of the real and generated images is calculated. These differences represent the perceptual dissimilarity between the real and generated images. The underlying intuition is that if the generated image is perceptually similar to the real image, their feature maps would also be very similar, leading to smaller differences. This way, the perceptual loss as measured by LPIPS incorporates more abstract visual characteristics and better aligns with human visual perception, thus contributing to the generation of higher quality images with the VQGAN model.
    \end{itemize}

\item \textbf{Additional Losses and Operations:}
    \begin{itemize}
    \item \textbf{Combined Perceptual and Reconstruction Loss:} After the perceptual loss is calculated using the LPIPS metric, it's combined with the L1 reconstruction loss calculated from the absolute difference between the original and the decoded images. These two losses are weighted by factors called as perceptual loss factor and rec loss factor respectively, before being added to form the perceptual rec loss. This loss is then averaged across all images in the batch. The rationale behind this operation is to strike a balance between perceptual similarity (appearance to the human eye) and raw pixel-level similarity.

    \item \textbf{Adversarial Loss:} The gan loss represents the adversarial loss for the generator, calculated as the negative mean of the fake outputs from the discriminator . The negative sign indicates that the generator's goal is to fool the discriminator into thinking the generated images are real.

    \item \textbf{Lambda Calculation:} The \texttt{calculate\_lambda} method in the code is used to dynamically adjust the balance between the VQGAN's perceptual reconstruction loss and the adversarial loss, depending on their magnitudes.

    \item \textbf{Final Loss:} The final loss is the sum of the perceptual reconstruction loss, the commitment and codebook losses, and the generator's adversarial loss, weighted by the discriminator factor and the dynamically calculated lambda. This combination of losses helps the VQGAN model to not only reconstruct images well, but also produce visually pleasing and diverse images that can fool the discriminator.
    \end{itemize}
\end{itemize}

\subsection{2d Positional Encoding}
The use of 2D positional encodings with the VQGAN architecture was motivated by the Neural Radiance Fields (NeRF) method, which uses positional encodings to synthesize novel views of complex 3D scenes from a sparse set of 2D images. Positional encodings in NeRF are essential as they represent high-frequency details like texture and lighting effects. In our VQGAN implementation, sine and cosine functions are applied alternately to different dimensions of a tensor, providing a smooth and continuous encoding of positions within 2D space. This alternating usage helps preserve relative positional information across various scales. Furthermore, a dropout layer enables the model to learn to depend on these encodings to an appropriate extent, introducing an additional regularization mechanism.

In the context of VQGAN, introducing 2D positional encodings could be highly beneficial. Images are inherently spatial structures, with the relationship between pixels in both horizontal and vertical dimensions carrying significant information. By integrating these encodings, the model can potentially learn richer representations of the images. Positional information enhances understanding of geometric patterns, alignment, and spatial hierarchies within the visual data. Therefore, incorporating 2D positional encoding into VQGAN may lead to improved performance and more nuanced image reconstructions.

\subsection{Image Reconstruction using Principal Component Analysis (PCA)}
\begin{itemize}
\item \textbf{Preprocessing and Channel Separation:}
The preprocessing step involves breaking down the images into their individual color channels, such as Red, Green, and Blue. This separation treats each color channel as an independent dataset, allowing for a more granular analysis of the color structure within the channel.
\item \textbf{Channel-Wise Analysis:}
For each color channel of every training image, PCA is applied. The pixels within a channel are treated as features in a multidimensional space. In this space, the principal components capture the most significant variations or patterns within the images.
\item \textbf{Identifying Principal Components:}
The principal components are vectors that highlight the directions where the variance is most significant among the images. By focusing on these directions, the essential information within the data can be summarized.
\item \textbf{Resulting PCA Models:}
The resulting PCA models for each channel, including the vectors and variance explained by each principal component, are stored. These models will be used for validation and reconstruction processes, capturing the underlying correlations among the pixels.
\item \textbf{Dimensionality Reduction:}
Validation images are transformed using the saved PCA models, reducing their complexity by keeping only the first n principal components. This compresses the information, while still preserving vital visual features.
\item \textbf{Reconstruction Process:}
The images are then reconstructed from the compressed representations. Though some information is lost during compression, most critical visual aspects are retained. The reconstructed images are saved to an output folder for further analysis.
\item \textbf{Proportional Analysis:}
The proportion of the total variance explained by the retained principal components is calculated. This offers an insight into the accuracy and efficiency of the reduced model in representing the original data.
\item \textbf{Visualizing Variance:}
Individual and cumulative variances are plotted to depict the contribution of each principal component and the overall variance explanation. Markers are used to highlight where 90\% and 95\% of the variance is explained, aiding in the selection of the optimal number of components to retain.
\end{itemize}

\section{Results and Discussions}
\label{sec:headings}



\subsection{Varying Latent Dimensions (64, 256, 512) with Fixed Codebook (1024) for 2700 Image Count}
\begin{figure}[htb]
\centering
\includegraphics[width=0.9\linewidth]{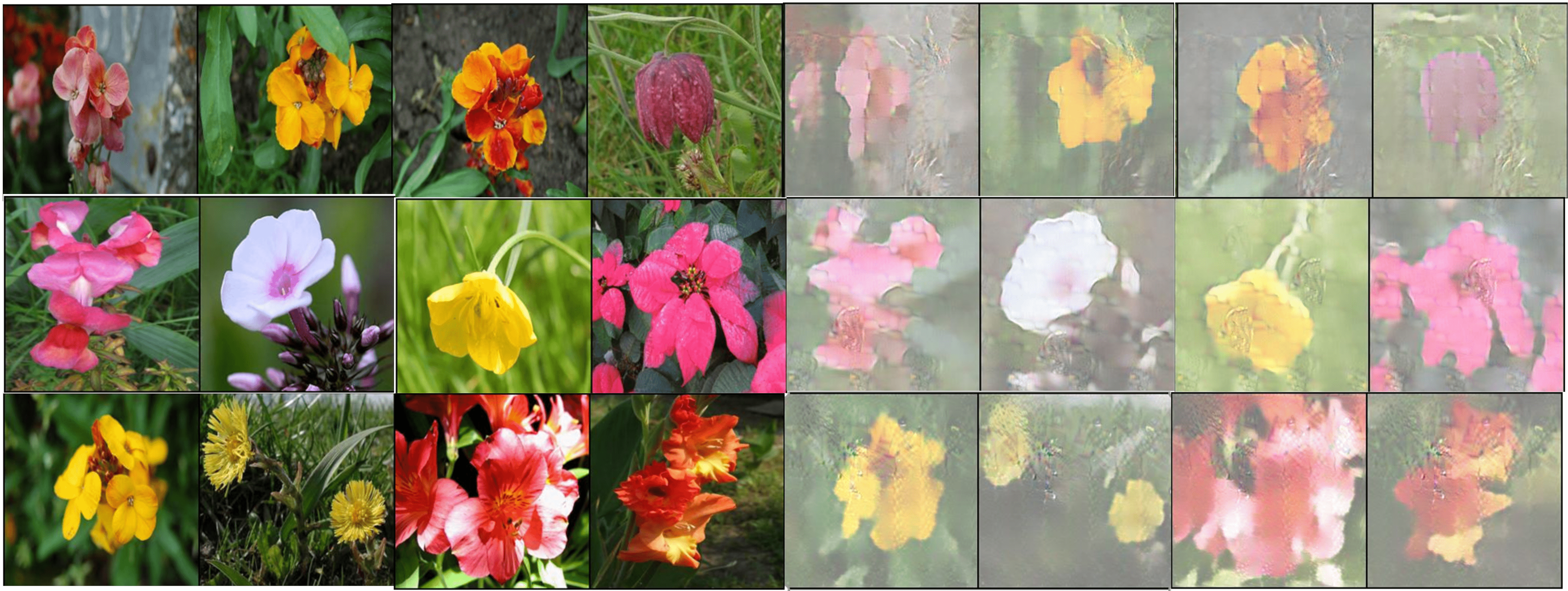}
\caption{Original Images (First Four), Reconstructed Images (Last Four) \textbf{(a)} \textbf{Row 1}: codebook size=1024, latent dimension size = 64, 300 epochs . \textbf{(b)} \textbf{Row 2}: codebook size=1024, latent dimension size = 256, 300 epochs. \textbf{(c)} \textbf{Row 3}: codebook size=1024, latent dimension size = 512, 300 epochs. (refer Figure \ref{fig10}(a))}
\label{fig2}
\end{figure}

The results from the experiments with a fixed codebook vector at 1024 and varying latent dimensions of 64, 256, and 512 present intriguing patterns. The presence of square artifacts in images with a latent dimension of 64 is particularly notable. These square artifacts could be manifestations of quantization error, where the model oversimplifies the encoded information, causing a significant loss in detail. It's a common phenomenon in models that struggle to capture the higher-order statistics of an image, often signifying a reduced ability to represent complex structures within the images. The square artifacts may indicate a kind of compression noise, where the model's lower latent dimension restricts the available information channels, leading to a patterned loss in detail.\\
In contrast, increasing the latent dimension to 256 showed improvements in some areas, such as the visibility of the flower's stem in Figure ~\ref{fig2}(b). This suggests that the latent dimension of 256 provides a more appropriate complexity level for this specific task. By expanding the latent dimension to 256, the model seems to capture essential image details better without introducing new challenges. It seems to offer a level of complexity that allows the encodings to represent the main features of the images, yet not adequately.
Yet, further increasing the latent dimension to 512 seemed to reintroduce challenges. Artifacts returned, and the reconstruction remained incomplete. This might be indicative of a more complex underlying problem where an over-increased latent dimension complicates the model's training. More complexity in the latent space can create a more rugged loss landscape, where the model finds it difficult to converge to an optimal solution.\\
The analysis of the VQ loss, as depicted in Figure ~\ref{fig10}(a), offers additional insights. With both training and validation loss not converging to 0, it's evident that 300 epochs were not sufficient for the model to learn the complexities of 2700 images with 1024 codebook vectors. The persistence of the loss near 0.03 for training and 0.05 for validation across epochs suggests an underlying inadequacy in the training process, perhaps pointing to the need for further hyperparameter tuning or a more sophisticated loss function.
Interestingly, the validation loss was lower for the runs with a latent dimension of 256, corroborating the qualitative observations. This points to a potentially optimal latent dimensionality for this particular task and dataset, further emphasizing the nuanced trade-offs between latent dimension, codebook size, and image quality.

\subsection{Varying Codebook Sizes (512, 1024, 2048) for 2700 Image Count with Fixed Latent Dimension (256)}
\begin{figure}[htb]
\centering
\includegraphics[width=0.9\linewidth]{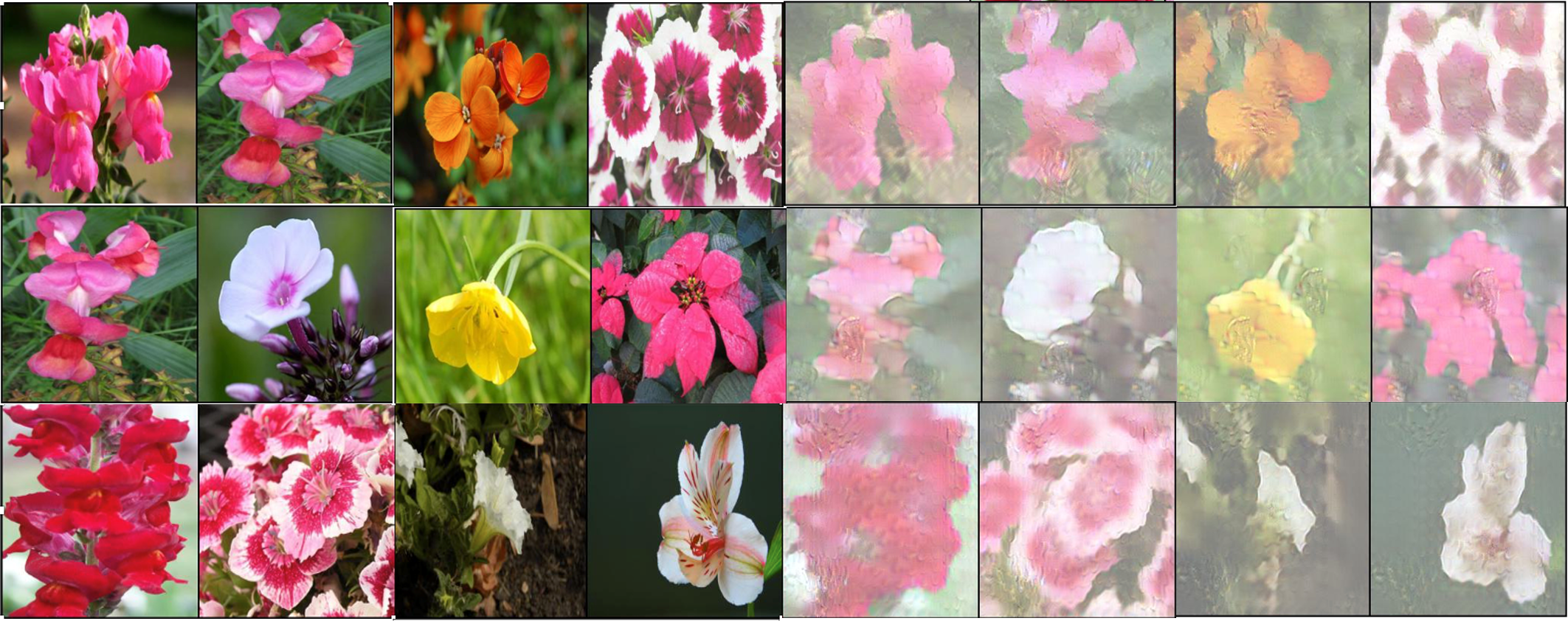}
\caption{Original Images (First Four), Reconstructed Images (Last Four) \textbf{(a)} \textbf{Row 1}: codebook size=512, latent dimension size = 256, 300 epochs. \textbf{(b)} \textbf{Row 2}: codebook size=1024, latent dimension size = 256, 300 epochs. \textbf{(c)} \textbf{Row 3}: codebook size=2048, latent dimension size = 256, 300 epochs.(refer Figure \ref{fig10}(b))}
\label{fig3}
\end{figure}

\begin{figure}[htb]
\centering

\includegraphics[width=0.9\linewidth]{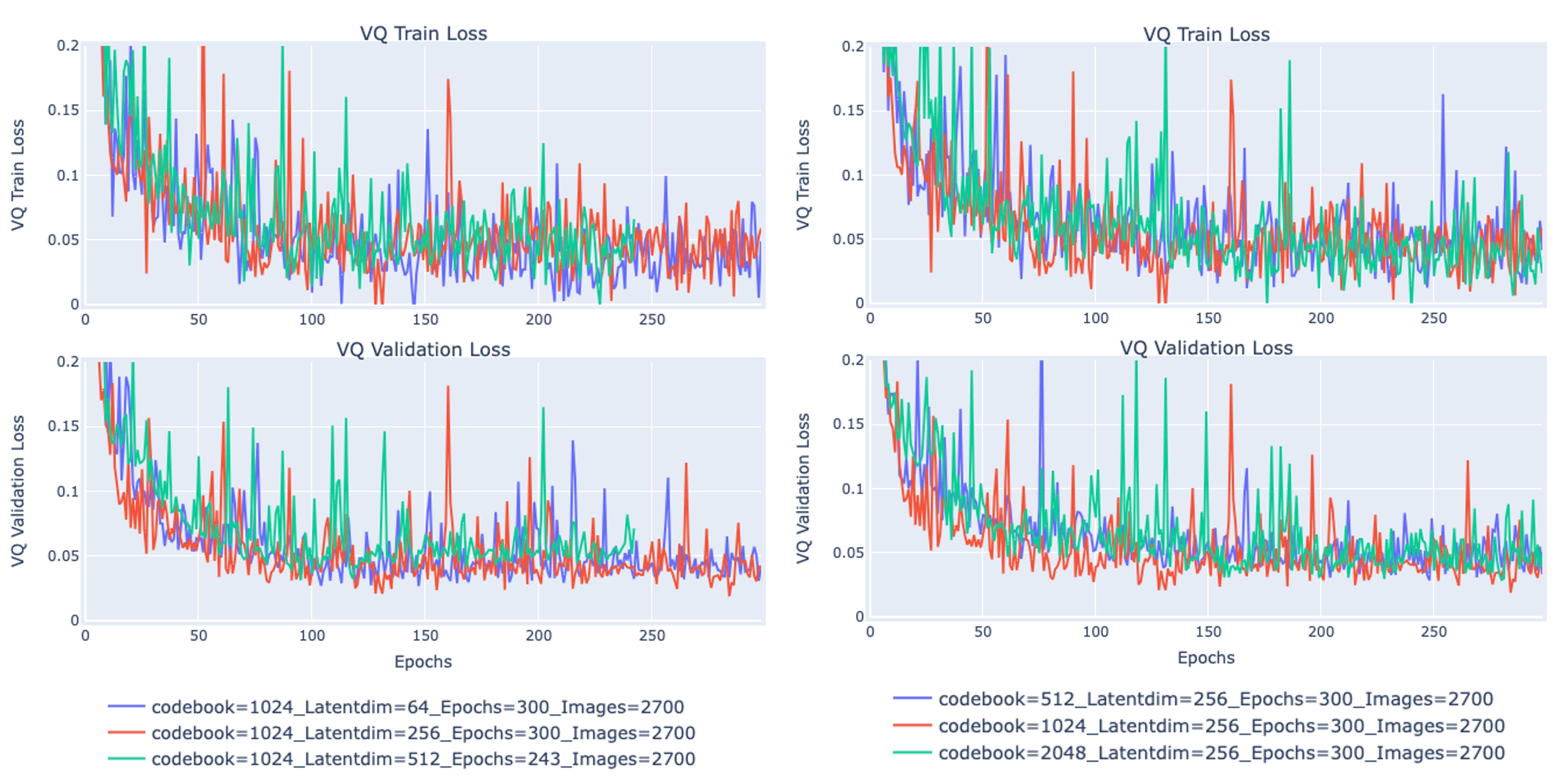}
\caption{\textbf{(a)} \textbf{Column 1}: codebook size = 1024, Latent dimensions $= [64, 256, 512]$, 2700 images. (refer Figure \ref{fig2})  \textbf{(b)} \textbf{Column 2}: codebook sizes$=[512, 1024, 2048]$, latent dimension size = 256, 2700 images. (refer Figure \ref{fig3})}
\label{fig10} 
\end{figure}

The different codebook sizes of 512, 1024, and 2048, when used with a latent dimension of 256, revealed diverse effects on the image quality Figure ~\ref{fig3}. For the codebook of size 512, an interesting observation was the appearance of spiral artifacts, especially in the lower portion of the images. This type of artifact might be an indication of localized information loss within the lower image regions. It's possible that this part of the image space was underrepresented or poorly approximated within the chosen codebook size, leading to a twisted, spiral pattern in the reconstruction. This could also hint at underlying structural imbalances within the dataset or model architecture that emphasize the upper regions of the image at the expense of the lower regions. Codebook of size 1024 is relatively better in comparison to codebook of 512, but with squared artifacts which might highlight the model inability to fully understand the spatial correlations within the images, leading to blocky reconstruction. \\
When the codebook size was increased to 2048, the reconstruction appeared to improve in color fidelity and shape representation. However, new artifacts emerged, particularly in the foreground or in-focus objects. The prevalence of these artifacts in areas with distinct and meaningful information (such as flowers) might indicate a bias within the encoding process. It could signify a struggle in differentiating between primary objects of interest and the background, resulting in misrepresentations of the complex features that distinguish the foreground.\\
In Figure ~\ref{fig10}(b), the convergence of both training and validation losses around 0.05 provides another dimension of insight. This consistent settling might be interpreted as a plateau in the learning process. If the model was still capable of significant improvements, one would expect to see continued reduction in loss. The fact that the loss remains stable could indicate that the model has reached the limits of what it can learn with the given configuration. It also underscores the relative effectiveness of a codebook of 1024, which performed the best in validation. This outcome reinforces the idea that choosing the appropriate codebook size is a nuanced task, balancing the need for detail with the risk of introducing various artifacts.

\subsection{Impact of Larger Codebook Sizes (4096, 8192) for 2700 Image Count with Latent Dimension (512)}
\begin{figure}[htb]
\centering
\includegraphics[width=0.9\linewidth]{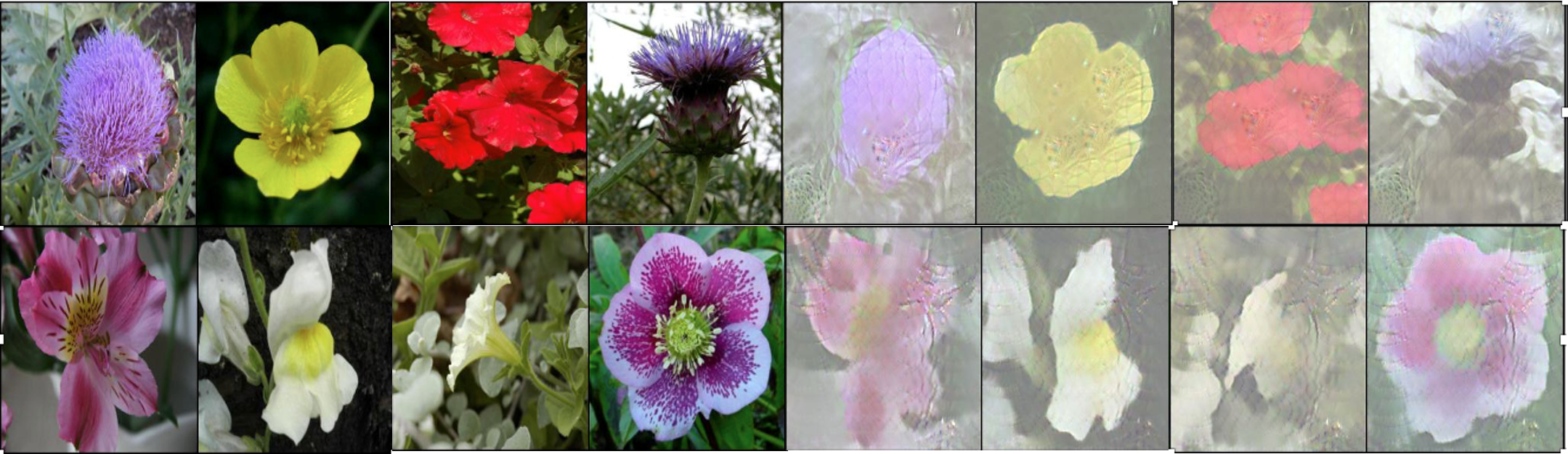}
\caption{Original Images (First Four), Reconstructed Images (Last Four) \textbf{(a)} \textbf{Row 1}: codebook size=4096, latent dimension size = 512, 300 epochs. \textbf{(b)} \textbf{Row 2}: codebook size=8192, latent dimension size = 512, 300 epochs.(refer Figure \ref{fig11}(b))}
\label{fig4}
\end{figure}

\begin{figure}[htb]
\centering

\includegraphics[width=0.9\linewidth]{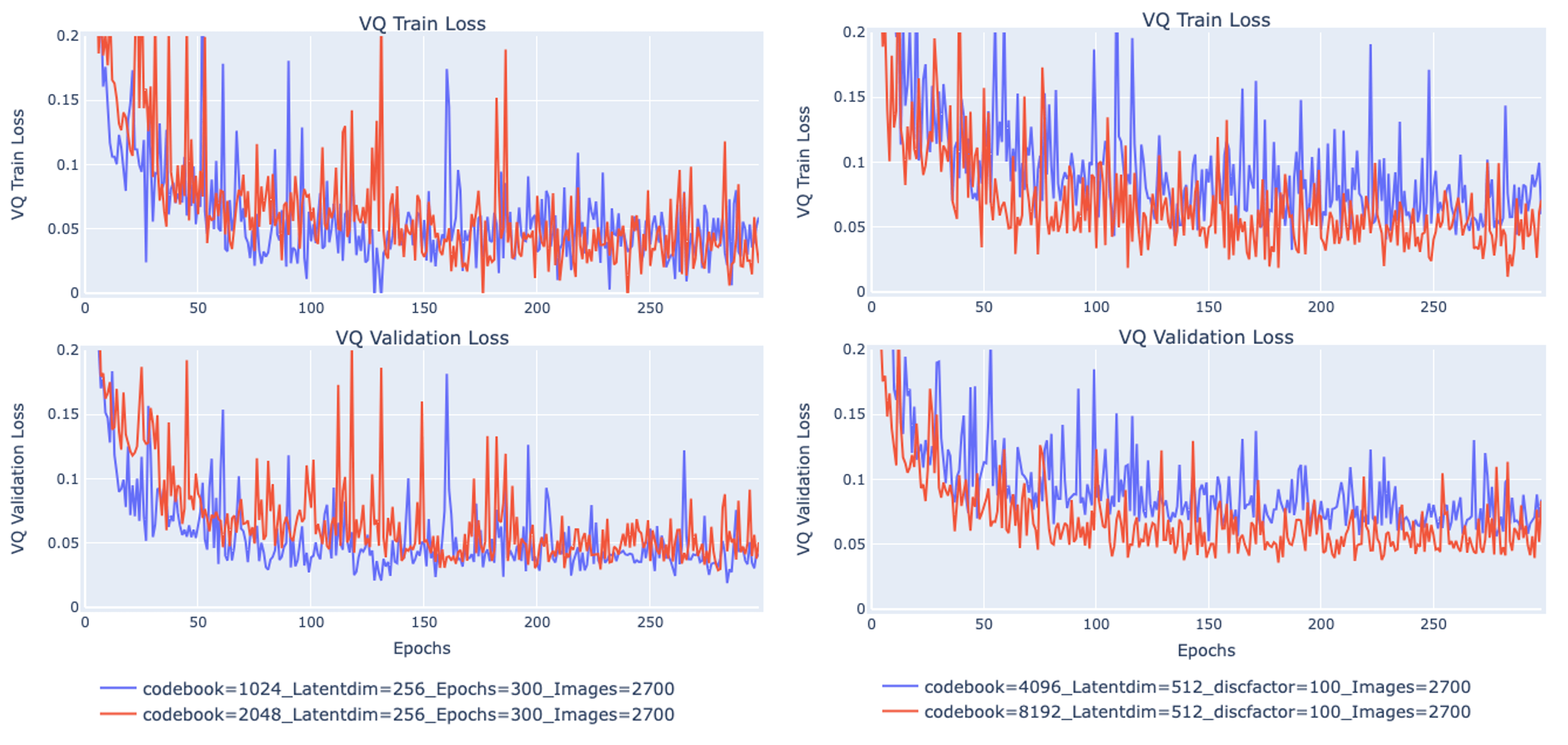}
\caption{\textbf{(a)} \textbf{Column 1}: codebook size $= [1024,2048]$, Latent dimensions = 256, 2700 images. (refer Figure \ref{fig3}) \textbf{(b)} \textbf{Column 2}: codebook size $= [4096,8192]$, latent dimension size = 512, 2700 images. (refer Figure \ref{fig4})}
\label{fig11} 
\end{figure}

The experiments with codebook vector sizes of 4096 and 8192 for Figure ~\ref{fig4}, and an increased latent dimension size of 512, yielded unexpected outcomes. While it was anticipated that larger codebook sizes would lead to superior performance compared to a codebook of 1024 with a latent dimension of 256, the reconstructions were plagued with artifacts. This puzzling result may suggest that merely scaling up these parameters does not linearly translate into better reconstructions. It could also imply that there is a more complex underlying relationship between codebook size, latent dimension, and image complexity that needs to be carefully balanced. For Figure ~\ref{fig4}(a), where codebook size was 4096, the images with high frequency and complex features, such as needle-like flowers in the foreground and abundant leaves in the background (Figure ~\ref{fig4}(a)), exhibited numerous artifacts. The dominant artifacts in the lower left corner might be indicative of a failure in capturing localized details within the image. It could be a manifestation of the model's inability to encode intricate spatial correlations within a specific region, possibly due to the relative scarcity of these features within the training set. For Figure ~\ref{fig4}(b), where the codebook size was 8192, the situation slightly improved with a fewer artifacts but still significant issues in color reconstruction. \\
The failure to regenerate specific details like the pink/reddish color of the flowers and the pollens might indicate that the larger codebook was better at capturing broad structures but still lacked the finesse needed for more nuanced aspects of the image. This could reflect a mismatch between the codebook granularity and the image's inherent complexity. The fact that the results were unsatisfactory even with large codebook sizes and higher latent dimensions relative to a small image count (2700 images) raises intriguing questions. One might expect that a larger codebook would allow more room for a detailed representation of the images, but the opposite seemed to occur. This disconnect suggests that something might indeed be off with the hyperparameter selection. The specific ratio between the codebook size and the number of images might not be optimal, leading to either over-complication or oversimplification in the learned representations.\\
In the case of a codebook size that is significantly larger than the number of images, it might create too much room for specialization, where the model focuses on representing minor variations and noise rather than general patterns. This could lead to a failure in capturing essential details, reflected in the training and validation losses settling around 0.05 ( Figure ~\ref{fig11}(b)). But if this was the case, then one would expect that training images to be modelled well, but that is not happening. Therefore, the failure to reconstruct even the training images suggests that this hypothesis may not explain the behaviour observed. \\
A more likely scenario could involve a mismatch between the granularity of the codebook and the inherent complexity within the images. If the codebook size is too large relative to the number of images, the model might struggle to find meaningful patterns that generalize well across the dataset. Instead of overfitting to the noise, the model could be underfitting, failing to capture both the main structures and the subtleties within the images. The codebook size might be introducing an additional complexity that the model is unable to utilize effectively given the particular dataset size and latent dimension. The hyperparameters might not be synergizing well, resulting in a failure to capture essential details in both training and validation sets. The losses settling around 0.05 could be indicative of the model's struggle to move beyond a suboptimal local minimum. Figure ~\ref{fig11}(b) showed that a codebook of 8192 was much better in comparison to a codebook of 4096, as both training and validation losses were lower. However, they settled around 0.05, further emphasizing that something is off with this selection of hyperparameters. 

\subsection{2D Positional Encodings: Impact with Reduced Image Count (185), Codebook (8192), and Latent Dimension (256)}
\begin{figure}[htb]
\centering
 \includegraphics[width=0.9\linewidth]{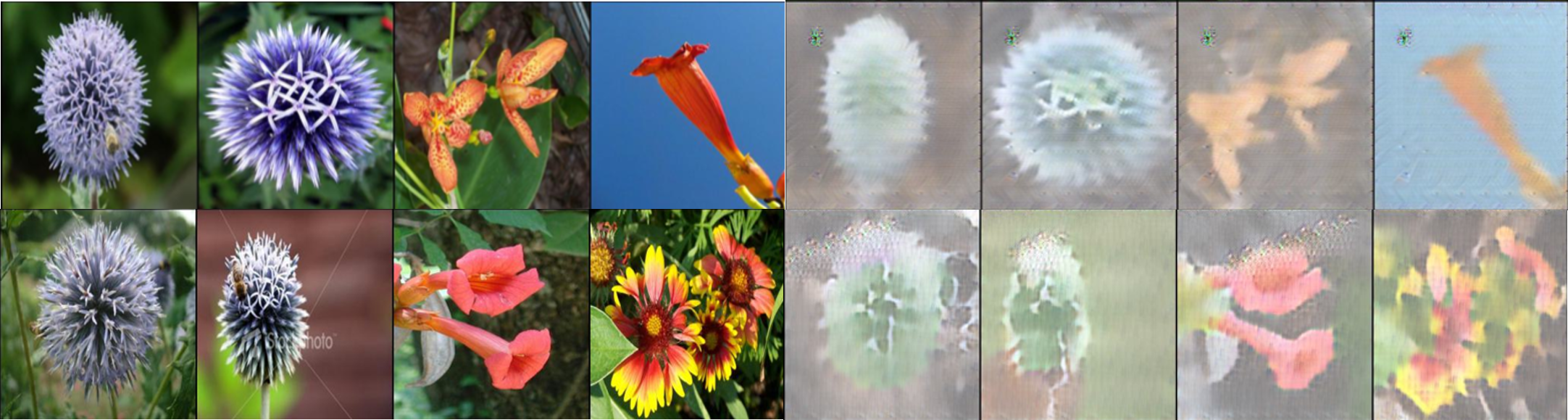}
\caption{Original Images (First Four), Reconstructed Images (Last Four) \textbf{(a)} \textbf{Row 1}: codebook size=8192, latent dimension size = 256, 185 images, without positional encoding. \textbf{(b)} \textbf{Row 2}: codebook size=8192, latent dimension size = 256, 185 images, with positional encoding. (refer Figure \ref{fig14}(a))}
\label{fig5}
\end{figure}

\begin{figure}[htb]
\centering

\includegraphics[width=0.9\linewidth]{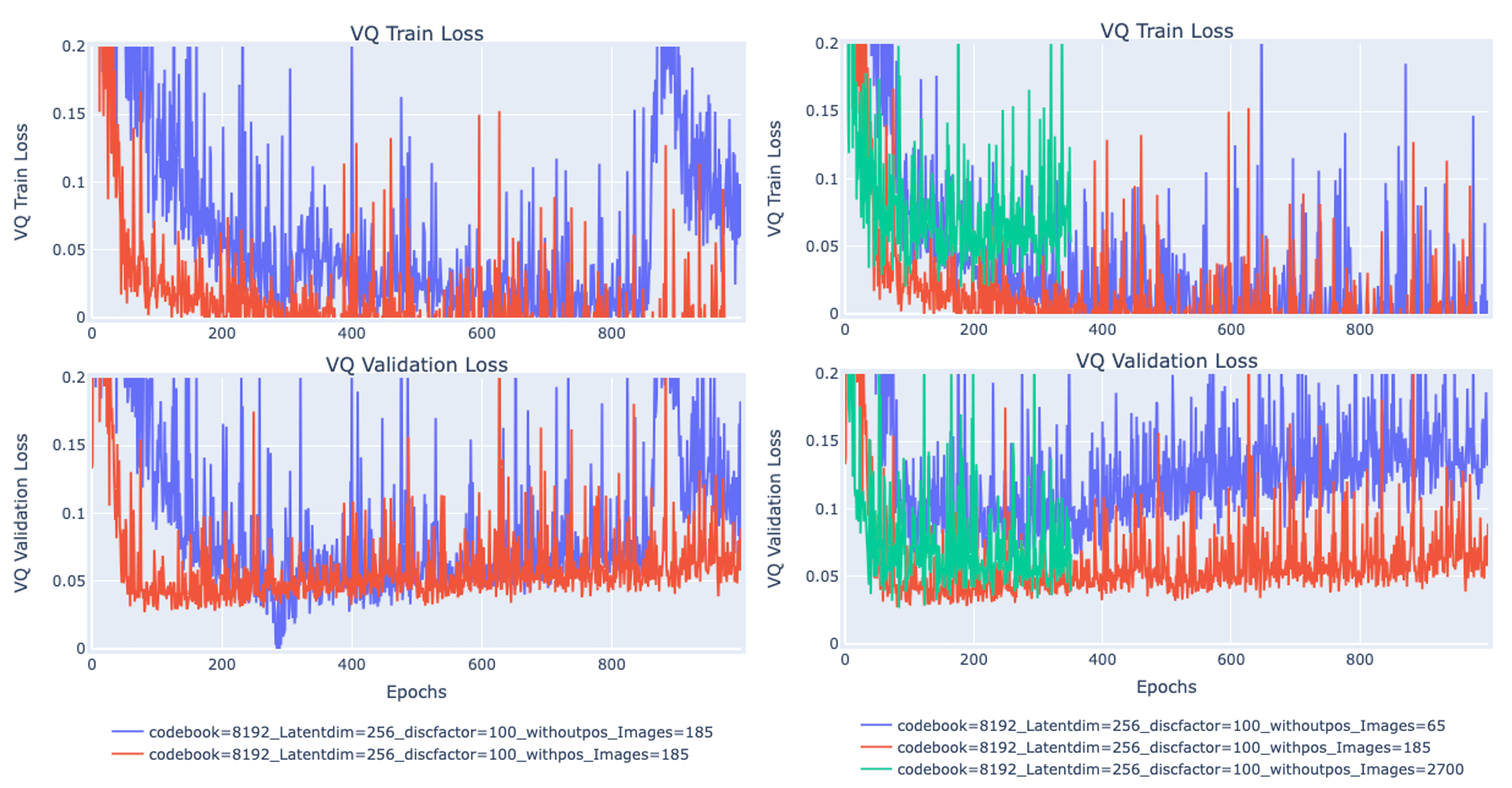}
\caption{\textbf{(a)} \textbf{Column 1}: codebook size = 8192, Latent dimensions = 256, 185 images, without and with positional encoding respectively. (refer Figure \ref{fig5}) \textbf{(b)} \textbf{Column 2}: codebook size = 8192, latent dimension size = 256, image sizes $=[65, 185, 2700]$, without, with and without positional encoding respectively. (refer Figure \ref{fig8} \& Figure \ref{fig5} \& Figure \ref{fig6})}
\label{fig14} 
\end{figure}

In pursuit of optimizing the model's performance, substantial alterations were made to both the dataset and the model. This process was driven by the observed high-frequency content in the images, which exhibited intricate variations in features such as pollen grains, leaves, and color changes. The first strategy was to reduce the number of images (we took 185 images for training). This was done to specifically train the model better. With lesser data, model indeed trained better, and the artifacts appearing before (square and spherical) were removed. However, the results were not entirely satisfactory. The features remained somewhat unclear, and the colors appeared unstable.\\
Recognizing the complex nature of the images, the next significant change introduced was the implementation of 2D positional encodings into the model (Figure ~\ref{fig5}(b)). Sinusoidal functions were used to encode the positional information across the image's height and width, incorporating both sine and cosine terms for alternating dimensions.\\
The outcome of this addition was mixed. While the background leaves became more visible, reflecting a better spatial understanding by the model, the features appeared smeared and looked worse than without positional encodings. These effects were further analyzed through the training and validation losses (Figure ~\ref{fig14}(a)). Without positional encodings, instabilities in training were evidenced by a sudden loss increase around 800 epochs, followed by a decrease. With positional encodings, more stable loss curves were achieved, but signs of overfitting emerged as validation losses increased after 100 epochs.\\
In conclusion, the refined experiments with a reduced image count and the introduction of 2D positional encodings yielded complex outcomes. While certain artifacts were mitigated and elements of image representation improved (Figure ~\ref{fig5}), new challenges such as smeared features and overfitting were unveiled (Figure ~\ref{fig14}(a)). The results underscore the delicate interplay between dataset size, model architecture, and hyperparameter tuning, especially when handling images with intricate and high-frequency content. Future efforts may further explore the nuanced effects of positional encodings and other encoding strategies, striving for an even more balanced and effective representation.

\subsection{2D Positional Encodings: Impact with Larger Image Count (2700), Codebook (8192), and Latent Dimension (256)}
\begin{figure}[htb]
\centering
\includegraphics[width=0.9\linewidth]{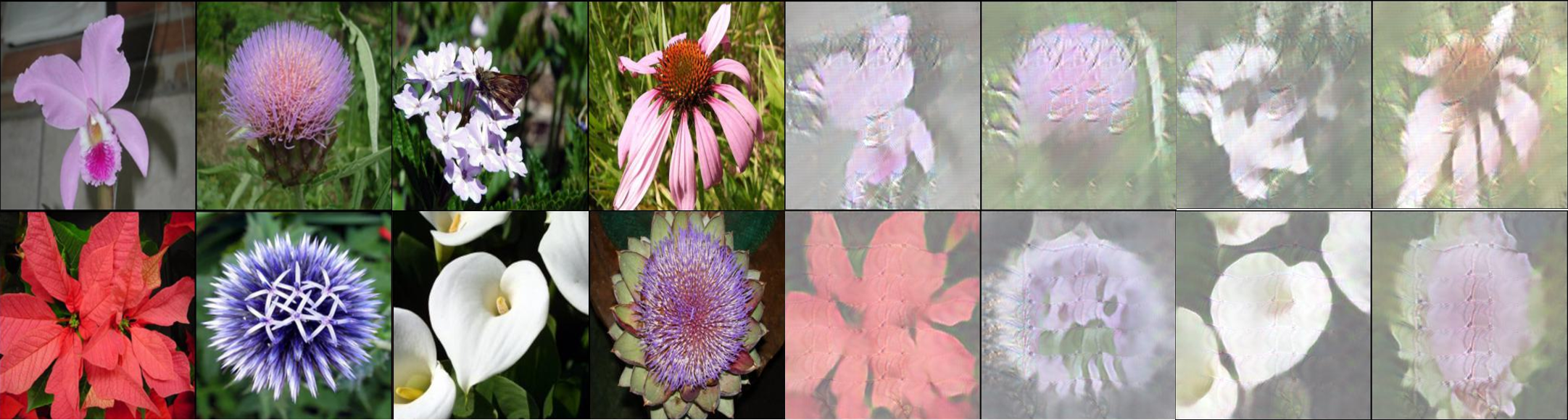}
\caption{Original Images (First Four), Reconstructed Images (Last Four) \textbf{(a)} \textbf{Row 1}: codebook size=8192, latent dimension size = 256, 2700 images, without positional encoding. \textbf{(b)} \textbf{Row 2}: codebook size=8192, latent dimension size = 256, 2700 images, with positional encoding. (refer Figure \ref{fig12}(a))}
\label{fig6}
\end{figure}

\begin{figure}[htb]
\centering

\includegraphics[width=0.9\linewidth]{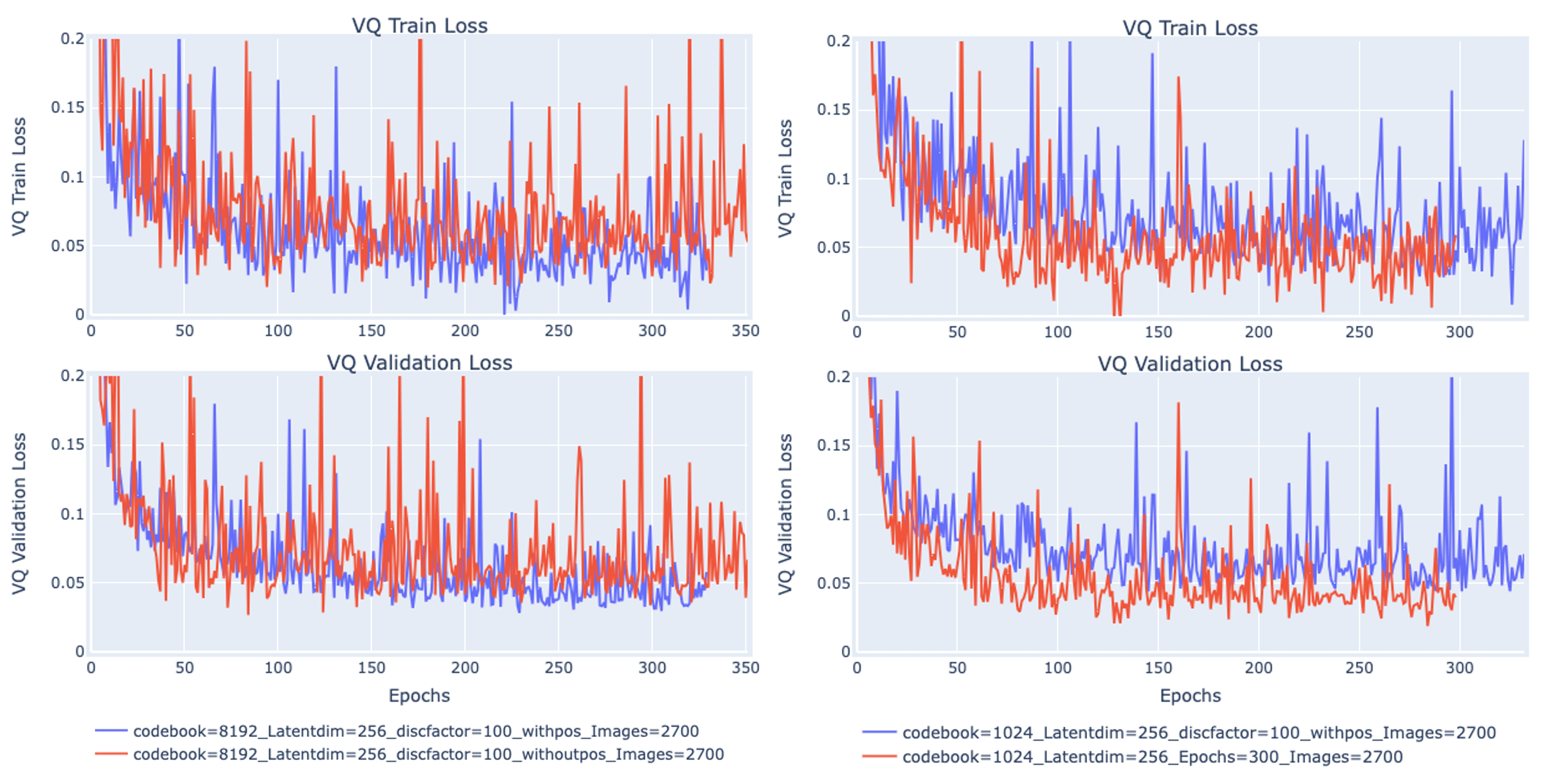}
\caption{\textbf{(a)} \textbf{Column 1}: codebook size = 8192, Latent dimensions = 256, 2700 images, with and without positional encoding. (refer Figure \ref{fig6}) \textbf{(b)} \textbf{Column 2}: codebook size = 1024, latent dimension size = 256, 2700 images, with and without positional encoding. (refer Figure \ref{fig7})}
\label{fig12} 
\end{figure}

In our continued exploration of model performance, we conducted experiments employing a codebook size of 8192 and a latent dimension of 256, utilizing 2700 images. This particular experiment was designed to analyze the effects of 2D positional encodings, both with and without their inclusion, as depicted in Figure ~\ref{fig6} \\
The immediate observation was a significant reduction in artifacts within the images when positional encodings were present in the system. This was a clear manifestation of the benefits of employing encodings to handle spatial correlations, especially in complex scenes with variations in features like pollen grains, leaves, and color changes. The introduction of positional encodings seemed to provide the model with a more nuanced ability to understand the spatial dependencies within the images, reducing the visual noise and artifacts that had plagued previous iterations of the model. \\
However, it was important to note that the improvements were not universal. When the images contained more intricate and rapidly changing features, artifacts were still present, even with the positional encodings. In these instances, the colors appeared smeared and less defined. While the positional encodings managed to handle simpler spatial correlations successfully, the more complex aspects of the images remained challenging. This highlighted the limitation of the current encoding mechanism and pointed to the need for further refinement to deal with high-frequency changes in features. \\
Despite that, the overall visual appearance of the model's reconstructions with positional encodings was better for the expanded set of 2700 images. The images appeared coherent and well-defined, suggesting a positive impact of the positional encodings on the model's ability to capture the essential structures within the images.
The training and validation loss curves further validated these observations. From Figure ~\ref{fig12}(a), it was evident that the model with positional encodings achieved lower validation and training losses. This quantitative evidence supported the visual inspection, pointing to better feature learning and generalization with positional encodings.

\subsection{2D Positional Encodings: Impact with Larger Image Count (2700), Varying Codebooks (1024, 8192), and Latent Dimension (256)}
\begin{figure}[htb]
\centering
\includegraphics[width=0.9\linewidth]{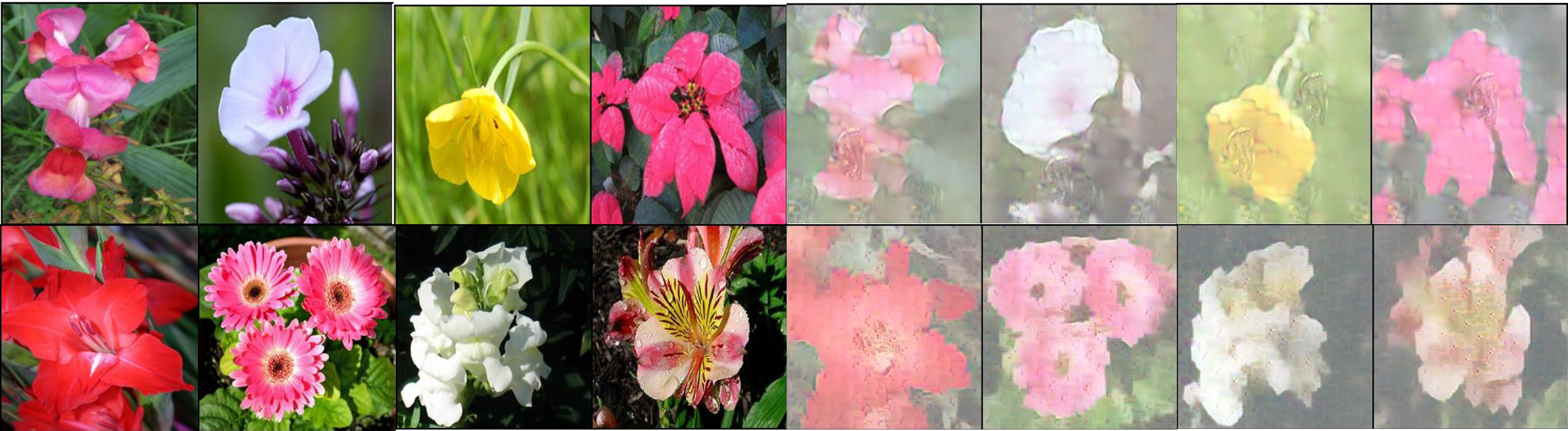}
\caption{Original Images (First Four), Reconstructed Images (Last Four) \textbf{(a)} \textbf{Row 1}: codebook size=1024, latent dimension size = 256, 2700 images, without positional encoding. \textbf{(b)} \textbf{Row 2}: codebook size=1024, latent dimension size = 256, 2700 images, with positional encoding. (Figure \ref{fig12}(b))}
\label{fig7}
\end{figure}

Next, we also tested the positional encodings with codebook of size 1024, and latent dimension of 256 (Figure ~\ref{fig7}(b)). The features appeared to be worse in comparison to the results of positional encodings with codebook of size 8192. There were more artifacts and the reconstruction was in a bad shape w.r.t to Figure ~\ref{fig6}(b). 
For the set of 2700 images, an intriguing observation emerged in the training and validation losses without positional encodings, as seen in Figure ~\ref{fig12}(b). Despite the visual deterioration, the loss metrics seemed to indicate better performance without positional encodings, as compared to the inclusion of them. However, both scenarios visually appeared to be unsatisfactory. This discrepancy between quantitative metrics and visual perception underscored the multifaceted nature of the problem and highlighted the potential limitations of relying solely on loss metrics for performance evaluation. \\
Figure ~\ref{fig13}(a) compares the results for codebook 8192, with position encodings and codebook 1024 without positional encodings. Intriguingly, the losses were nearby for both configurations but seemed to be stagnant at 0.05 for approximately 300 epochs. However, the visual inspection revealed a clear preference for the codebook 8192 with positional encoding (Figure ~\ref{fig6}(b) and Figure ~\ref{fig7}(a)), rendering more aesthetically pleasing results in comparison to the codebook 1024 without positional encodings. 

\subsection{2D Positional Encodings: Impact with Smaller Image Count (65), Codebook (8192), and Latent Dimension (256)}
\begin{figure}[htb]
\centering
\includegraphics[width=0.9\linewidth]{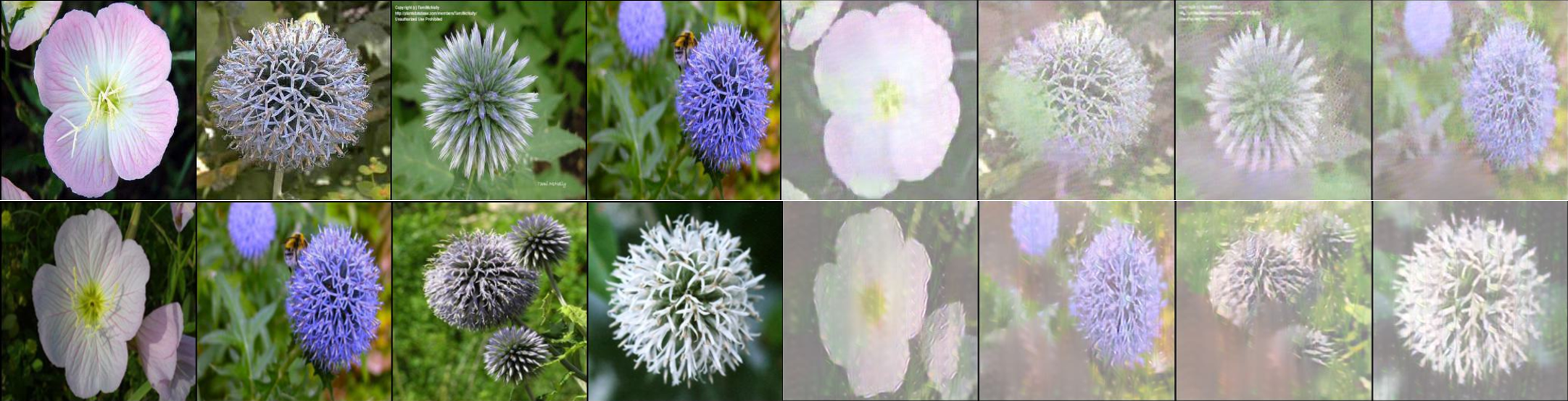}
\caption{Original Images (First Four), Reconstructed Images (Last Four) \textbf{(a)} \textbf{Row 1}: codebook size=8192, latent dimension size = 256, 65 images, without positional encoding. \textbf{(b)} \textbf{Row 2}: codebook size=8192, latent dimension size = 256, 65 images, with positional encoding. (Figure \ref{fig13}(b))}
\label{fig8}
\end{figure}

\begin{figure}[htb]
\centering

\includegraphics[width=0.9\linewidth]{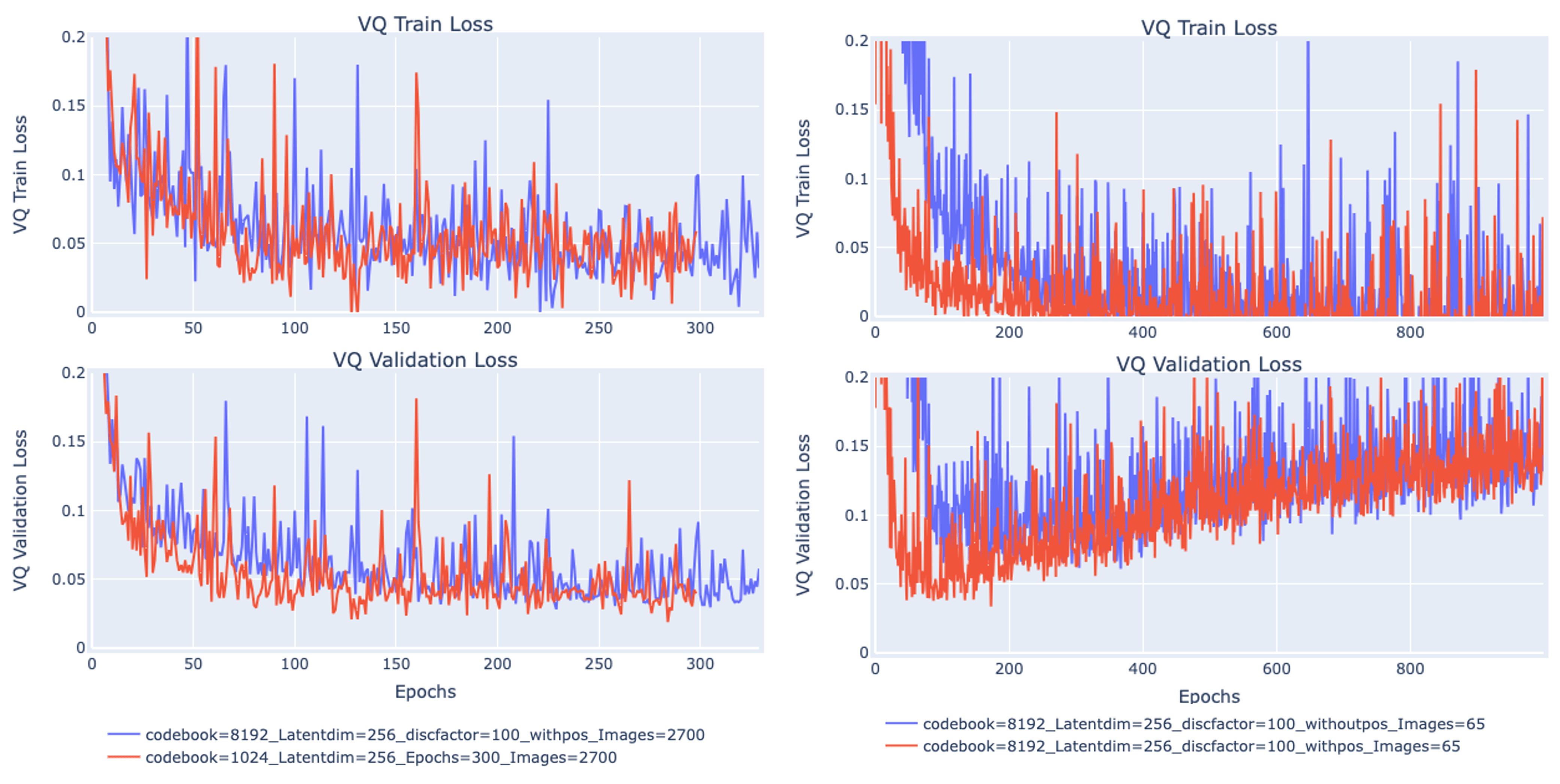}
\caption{\textbf{(a)} \textbf{Column 1}: codebook sizes $= [8192,1024]$, Latent dimensions = 256, 2700 images, with and without positional encoding respectively. (refer Figure \ref{fig6} \& \ref{fig7}) \textbf{(b)} \textbf{Column 2}: codebook size = 8192, latent dimension size = 256, 65 images, with and without positional encoding. (refer Figure \ref{fig8})}
\label{fig13} 
\end{figure}

In a further experiment involving a highly reduced dataset of just 65 images, we continued to investigate the effects of 2D positional encodings with a codebook vector size of 8192 and a latent dimension of 256 (Figure ~\ref{fig8}). By narrowing down the number of images and extending the training epochs up to 5800, this test provided a more focused examination of the positional encoding's impact on specific images. \\
The visual representation when positional encodings were included appeared to be considerably more aligned with the original images. This underscores the ability of positional encodings to capture spatial correlations more effectively, especially in a more constrained dataset. On the contrary, without positional encodings, certain images (specifically images 2 and 3) exhibited green color appearances in parts of the flowers. This manifestation suggested reconstruction challenges, hinting at the model's struggle to fully adapt to the nuances of these specific images without positional encodings.
In image 2, while most parts rendered the features accurately, a small portion displayed smeared features with a green hue. Image 3 was in better shape, yet a hint of green color was still present. These anomalies may signify the model's limitations in capturing high-frequency details without the spatial cues provided by positional encodings. \\
Both scenarios, with and without positional encodings, lacked background information. This missing background could be attributed to the model prioritizing more prominent features over subtler background details, or it could indicate the model’s inability to represent more complex and varying color patterns. This also highlights a challenge bin maintaining fullrichness of color and deatil in learning process. 
An interesting observation from the loss metrics was that both the training losses appeared to converge towards zero (Figure ~\ref{fig13}(b)). However, signs of overfitting emerged as the validation loss started to increase after 150 epochs for both scenarios. This phenomenon could indicate that while the model was becoming proficient in learning the training data, it may have been too specialized, losing its ability to generalize well to unseen data.

\subsection{2D Positional Encodings and Model Size Reduction, Smaller Image Count (65), Codebook (8192), and Latent Dimension (256)}
\begin{figure}[htb]
\centering
\includegraphics[width=0.9\linewidth]{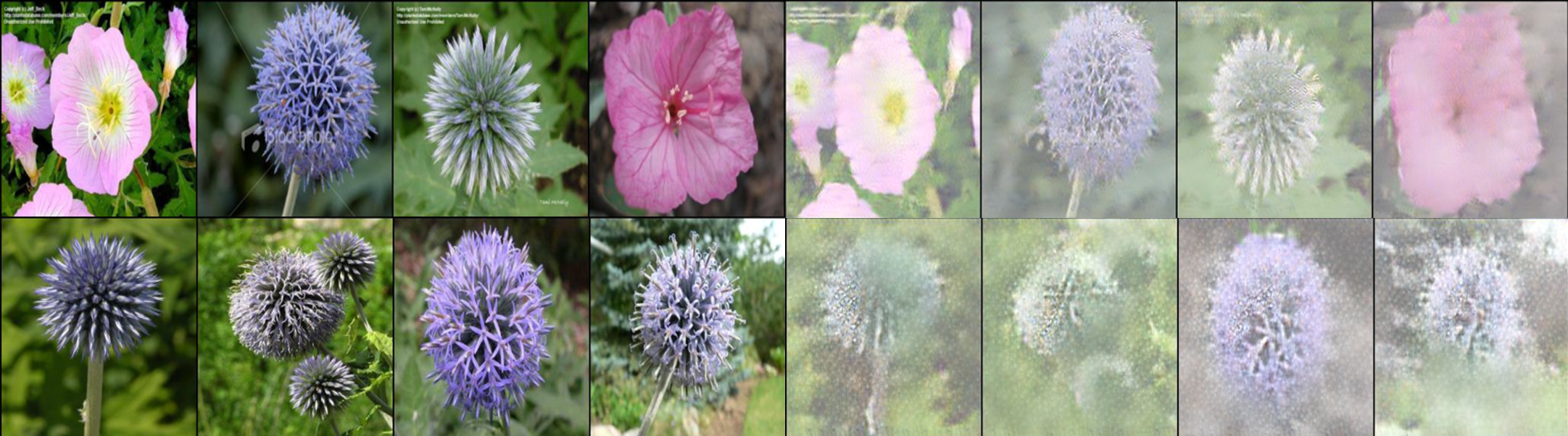}
\caption{\textbf{Smaller Network}, Original Images (First Four), Reconstructed Images (Last Four) \textbf{(a)} \textbf{Row 1}: codebook size=8192, latent dimension size = 256, 65 images, without positional encoding. \textbf{(b)} \textbf{Row 2}: codebook size=8192, latent dimension size = 256, 65 images, with positional encoding. (Figure \ref{fig15}(b))}
\label{fig9}
\end{figure}

\begin{figure}[htb]
\centering
\includegraphics[width=0.9\linewidth]{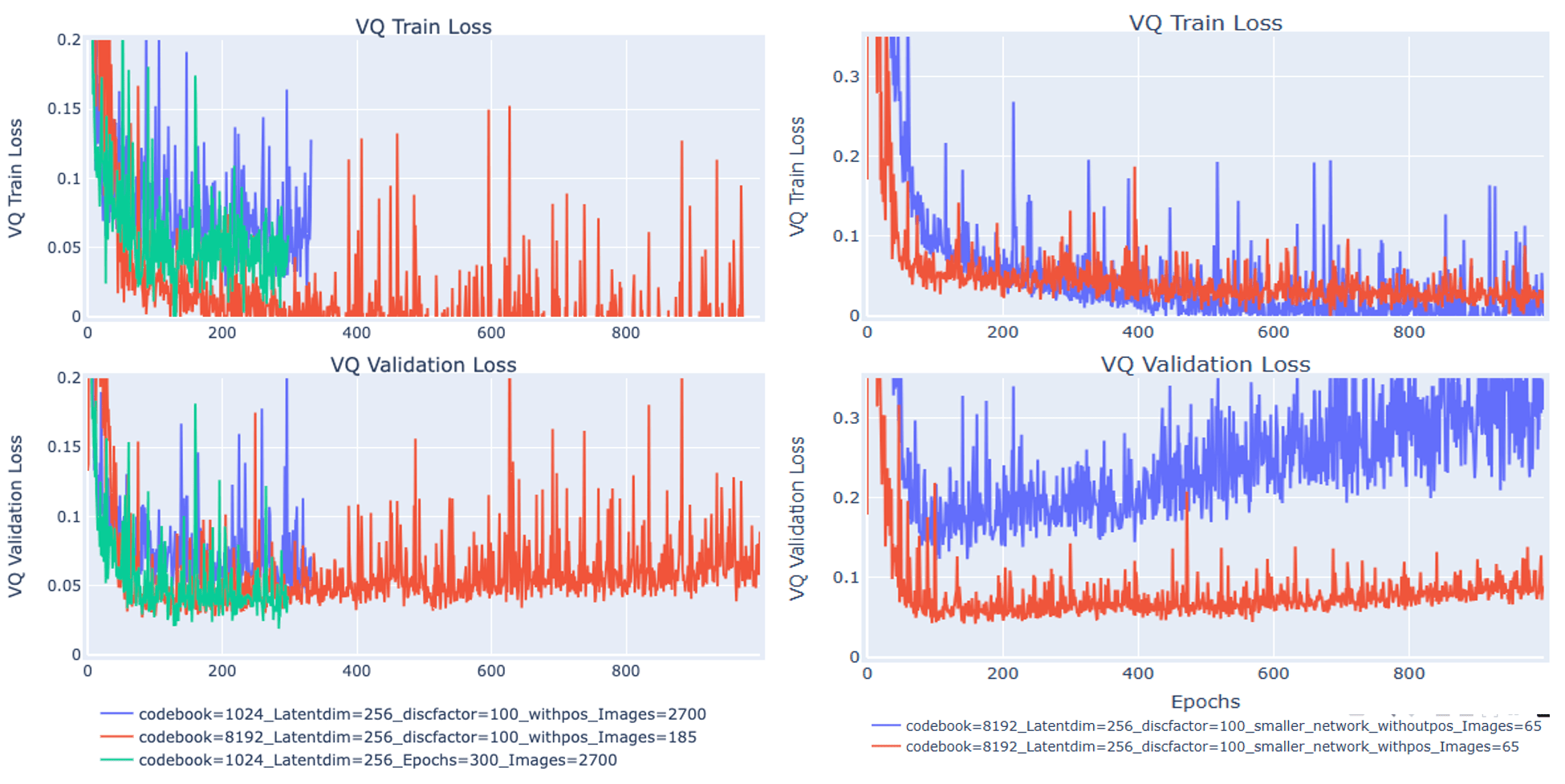}
\caption{\textbf{(a)} \textbf{Column 1}: codebook sizes = [1024, 8192, 1024], Latent dimensions = 256, image sizes $=[2700, 185, 2700]$, with, with and without positional encoding respectively. (refer Figure \ref{fig7} \& Figure \ref{fig5}) \textbf{(b)} \textbf{Column 2}: \textbf{Smaller  network}, codebook size = 8192, latent dimension size = 256, image size = 65, without and with positional encoding. (refer Figure \ref{fig9})}
\label{fig15} 
\end{figure}

In a subsequent experiment, we attempted to refine the model by reducing the number of layers in both the encoder and decoder, aiming to simplify its structure. The model was tested with and without positional encodings, using a codebook of 8192 and a latent dimension of 256 (Figure ~\ref{fig9}). Intriguingly, the results were more favorable for the model without positional encodings. The reconstructed features were much richer, and the background information was better represented compared to Figure ~\ref{fig8}. This finding suggests that a simpler model, with fewer layers in the encoder and decoder, was more adept at capturing the background information than a more complex one. \\
However, the introduction of positional encodings to this simpler model resulted in images that appeared dotted and unsmooth, with smeared features (Figure ~\ref{fig9}). In the losses, it was also observed that for smaller network, with positional encoding, the model was not overfitting, which was not the case when positional encodings were not employed (Figure ~\ref{fig15}(b)). Despite these improvements, we were still unable to eliminate the whitish nature of the reconstructed image. The persistence of this whitish appearance may indicate underlying challenges in accurately reproducing the original color spectrum or a tendency of the model to gravitate towards a neutral color palette when faced with complex visual information. It emphasizes the delicate trade-offs between complexity, feature representation, and visual fidelity that must be managed in model design.

\subsection{Comparative Analysis of Smaller Dataset (65 Images) and Larger Dataset (2700 Images) using Principal Component Analysis}
We conducted an experiment using Principal Component Analysis (PCA) for image reconstruction with both smaller (65 images) and larger (2700 images) datasets, utilizing 50 principal components for representation. The explained variance for the smaller dataset was approximately \(98.19\%\), while for the larger one it was roughly \(98.35\%\) (Figure ~\ref{fig16}). 

\begin{figure}[htb]
\centering

\includegraphics[width=0.9\linewidth]{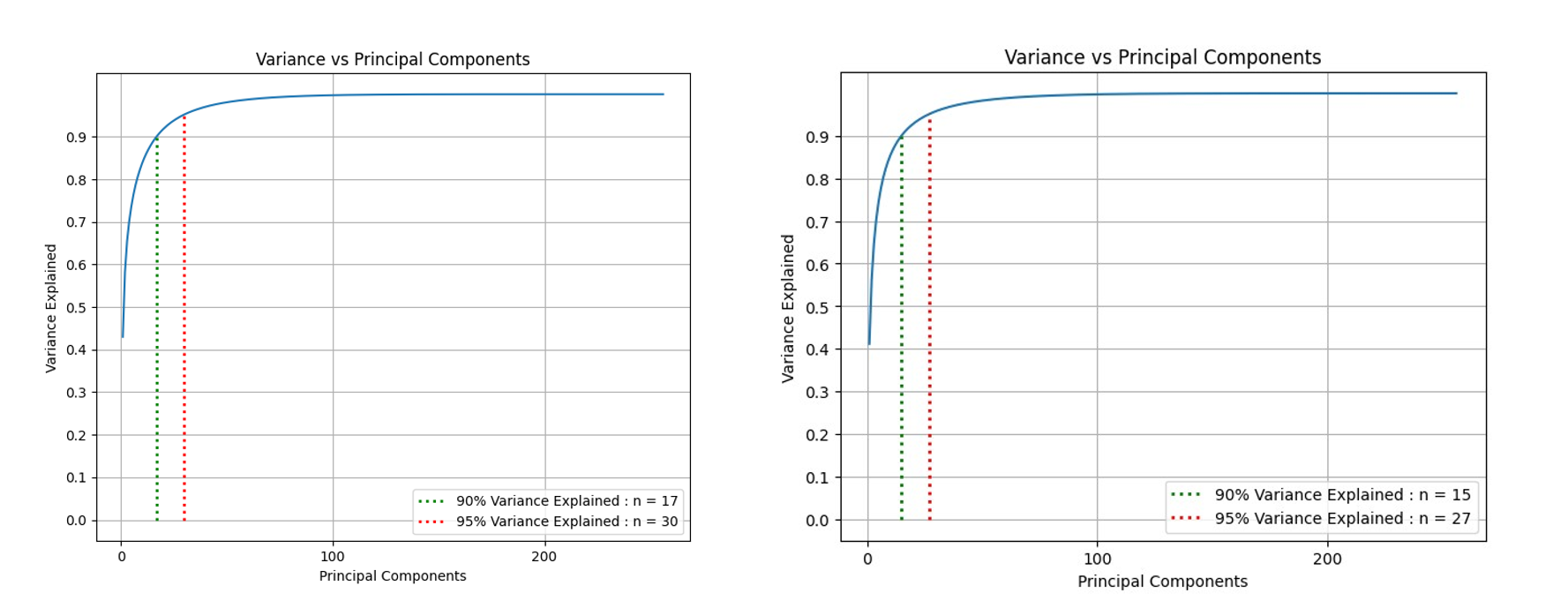}
\caption{\textbf{(a)} \textbf{Column 1}: variance vs. principal components for 65 images. \textbf{(b)} \textbf{Column 2}: variance vs. principal components for 2700 images.}
\label{fig17} 
\end{figure}

\begin{figure}[htb]
\centering

\includegraphics[width=0.9\linewidth]{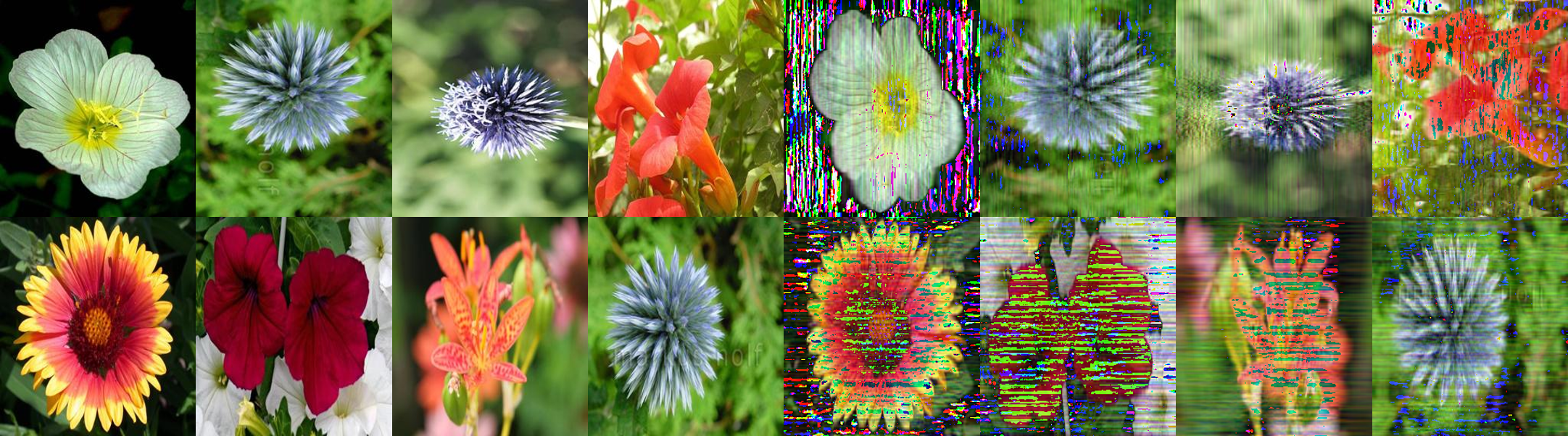}
\caption{Original Images (First Four), Reconstructed Images (Last Four) \textbf{(a)} \textbf{Row 1}: PCA reconstruction, 65 images. \textbf{(b)} \textbf{Row 2}: PCA reconstruction, 2700 images.}
\label{fig16} 
\end{figure}

However, visual inspections revealed a marked degradation in quality between the two datasets(Figure ~\ref{fig16}). In the 65-image dataset, the reconstructions showed anomalies like purplish lines in the background and bluish color distortions. For the 2700-image dataset, the performance was even worse, with greenish horizontal lines appearing in the foreground and failure in capturing essential details. Several factors may contribute to this unexpected result. The increased complexity and variability within the larger dataset might have overwhelmed PCA's ability to represent finer details. Since PCA relies on linear assumptions, it might have failed to handle the nonlinear structures and dependencies that become more pronounced with the increase in data complexity. The \(1.5\%-2\%\) unexplained variance might contain critical information affecting the visual quality, especially in a larger, more intricate dataset. Moreover, the choice of 50 components might have been insufficient for capturing nuanced variations in the larger dataset, despite sufficing for the smaller one. These observations highlight PCA's limitations in handling highly complex image data and emphasize that capturing a high percentage of variance does not guarantee accurate or visually pleasing reconstruction.

\section{Conclusion}
\begin{itemize}
    \item This investigation, though constrained by limited resources on a single A100 GPU for \(8\) hours, marks a significant advancement in understanding VQGAN behavior and design. With \(2700\) images, \(300\) epochs were reached, and with \(185\) and \(65\) images, we achieved \(1000\) epochs each, findings that shed light on the interplay between model sizes and image quantities.
    \item Importantly, the study is largely based on the effects on vector quantization loss, as GAN loss was not played or considered as a hyperparameter. This focus on vector quantization offers a specialized view and contributes to the nuanced understanding of loss functions in the context of VQGAN.
    \item The allocation of \(300\) epochs for \(2700\)-image dataset fell short in fully exploring varying codebook sizes \((512, 1024, 2048, 4096, 8192)\) and different latent dimensions \((64, 256, 512)\). This highlights a crucial area for further exploration in VQGAN and provides guidance for future researchers working on similar problems.
    \item A key finding was that reconstructions suffered more from artifacts with a larger dataset \(2700\) images compared to smaller ones \((185\) or \(65\)) images.
    \item We discovered that reconstructions with a lower codebook size \((1024)\) exhibited more artifacts than those using an \(8192\)-sized codebook with a fixed latent dimension of \(256\), an insight that can guide optimal codebook size selection in VQGAN design.
    \item The study emphasizes the superiority of deep learning models in capturing color features compared to PCA, underlining the potential of advanced models in this domain.
    \item The introduction of \(2\)D positional encoding in VQGAN, in combination with \(2700\) images, an \(8192\)-sized codebook, and latent dimensions of \(256\), brought about a significant reduction in artifacts, though some persisted. This finding demonstrates promising avenues for further refinement, particularly with an \(8192\) codebook as opposed to a \(1024\) size.
    \item Using \(185\) images, an \(8192\)-sized codebook, and latent dimension of \(256\), we observed the removal of specific artifacts but also early signs of overfitting, a critical observation for balancing clarity and overfitting in VQGAN models.
    \item By restricting the dataset to \(65\) images with a codebook of \(8192\) and latent dimension of \(256\), we managed to eliminate artifacts and align visual representations more closely to original images, though overfitting emerged. This provides vital lessons on dataset-size effects in VQGAN.
    \item Our study revealed that reduction in the size of encoder and decoder, lead to pleasing results as background was captured better. Interestingly, with smaller model, the overfitting was way lesser in model with positional encoding in comparison to when model size was bigger or when the model size was smaller but there were no positional encodings.
    \item Through exploring PCA, we found that even when \(50\) components explained over \(98\%\) of the variance for both smaller and larger datasets \(65\) and \(2700\) images, the quality was still inferior to deep learning models, particularly with the larger dataset, adding to the understanding of methodological selection in VQGAN.
\end{itemize}


\bibliographystyle{plainnat}
\bibliography{references}

\end{document}